\documentclass[11pt,a4paper]{article}
\usepackage[hyperref]{emnlp2020}
\usepackage{times}
\usepackage{latexsym}

\usepackage{microtype}
\usepackage{booktabs}

\usepackage{times}
\usepackage{latexsym}
\usepackage{url}
\usepackage{graphicx}
\usepackage{amsmath}
\usepackage{amssymb}
\usepackage{subfigure}
\usepackage{caption}
\usepackage{algorithm}
\usepackage{algorithmicx}
\usepackage{algpseudocode}
\usepackage{multicol}
\usepackage{multirow}
\usepackage{booktabs}
\usepackage{listings}
\usepackage{array}
\usepackage{bm}

\aclfinalcopy 

\title{Know What You Don't Need: \\ Single-Shot Meta-Pruning for Attention Heads}

 \author{Zhengyan Zhang$^{1,2,3}$, Fanchao Qi$^{1,2,3}$, Zhiyuan Liu$^{1,2,3\dagger}$, Qun Liu$^4$, Maosong Sun$^{1,2,3}$\\
        $^1$Department of Computer Science and Technology, Tsinghua University, Beijing, China  \\
         $^2$Institute for Artificial Intelligence, Tsinghua University, Beijing, China \\ 
         $^3$State Key Lab on Intelligent Technology and Systems, Tsinghua University, Beijing, China\\
         $^4$Huawei Noah's Ark Lab \\
         \texttt{\{zy-z19, qfc17\}@mails.tsinghua.edu.cn}\\
}

\date{}

\begin{document}
\maketitle
\begin{abstract}
Deep pre-trained Transformer models have achieved state-of-the-art results over a variety of natural language processing (NLP) tasks. By learning rich language knowledge with millions of parameters, these models are usually overparameterized and significantly increase the computational overhead in applications. It is intuitive to address this issue by model compression. In this work, we propose a method, called Single-Shot Meta-Pruning, to compress deep pre-trained Transformers before fine-tuning. Specifically, we focus on pruning unnecessary attention heads adaptively for different downstream tasks. To measure the informativeness of attention heads, we train our Single-Shot Meta-Pruner (SMP) with a meta-learning paradigm aiming to maintain the distribution of text representations after pruning. Compared with existing compression methods for pre-trained models, our method can reduce the overhead of both fine-tuning and inference. Experimental results show that our pruner can selectively prune 50\% of attention heads with little impact on the performance on downstream tasks and even provide better text representations. The source code will be released in the future.
\end{abstract}

{\let\thefootnote\relax\footnotetext{$^\dagger$ Corresponding author: Z.Liu (liuzy@tsinghua.edu.cn)}}

\section{Introduction}

Pre-trained language models (PLMs), such as BERT \citep{devlin2018bert}, XLNet \citep{yang2019xlnet} and RoBERTa \citep{liu2019roberta}, have achieved state-of-the-art results across a variety of natural language processing (NLP) tasks. 
To fully utilize large-scale unsupervised data during pre-training, PLMs are becoming larger and larger. 
For example, GPT-3 \citep{brown2020language} has 175 billion parameters. 
With the growing number of model parameters, the computational overhead, including memory and time, becomes tremendously heavy, which severely limits the application of PLMs to downstream NLP tasks.
Therefore, model compression for PLMs is increasingly important.

\begin{figure}
    \centering
    \includegraphics[width=\linewidth]{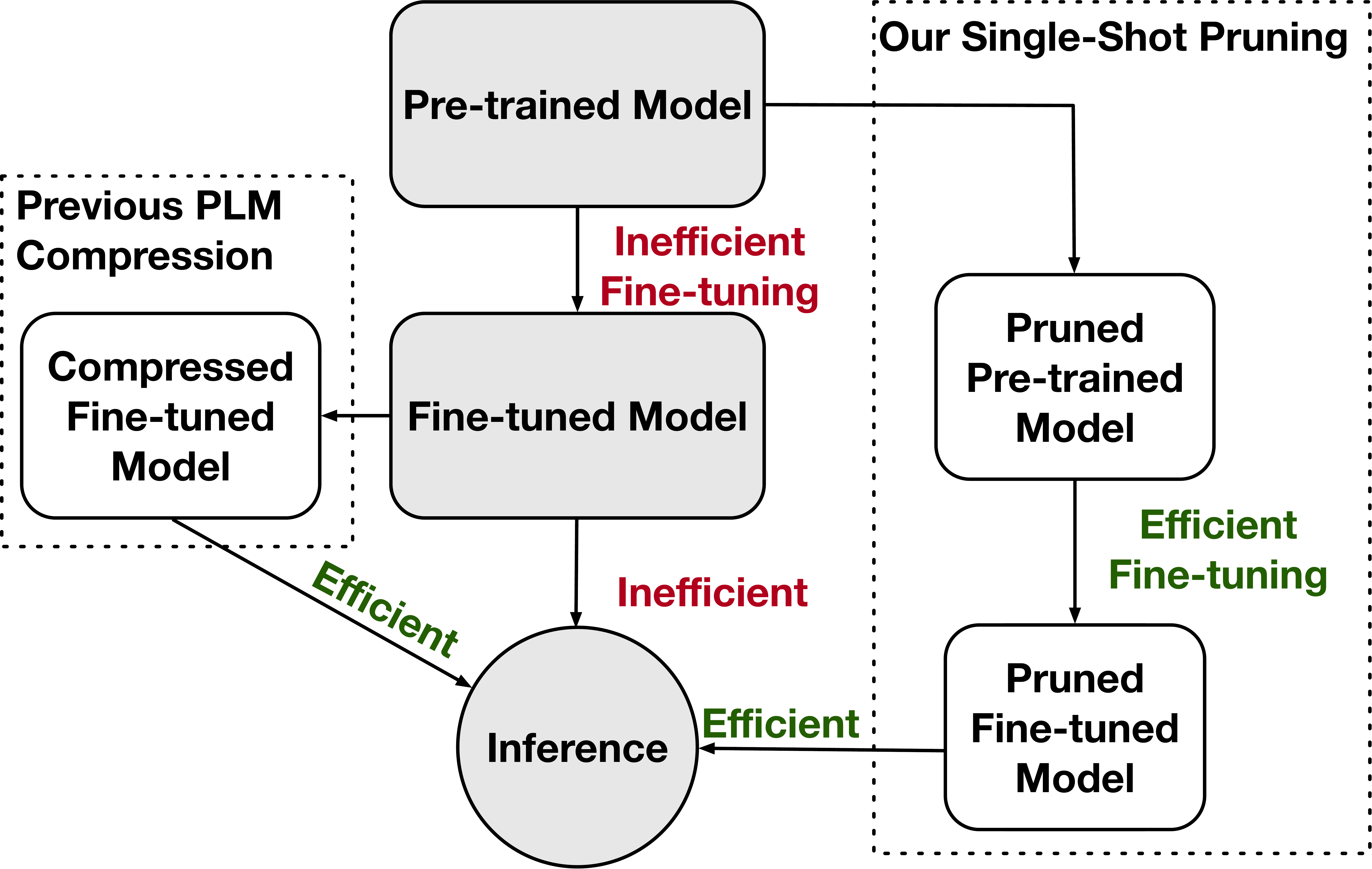}
    \captionsetup{font={normalsize}}
    \caption{An illustration of the three-step paradigm (gray blocks) adopted by most of PLMs and two types of model compression methods. Compared to previous work, our method makes both fine-tuning and inference efficient.}
    \label{fig:paradigm}
\end{figure}


Most PLMs are Transformer-based \citep{radford2018improving,devlin2018bert},
and they are utilized following a three-step paradigm: pre-training, fine-tuning and inference, as illustrated in Figure \ref{fig:paradigm}. 
Quite a few methods have been proposed to compress pre-trained Transformers during or after fine-tuning to reduce the computational overhead in the inference phase \citep{tang2019distilling,turc2019well,jiao2019tinybert,mccarley2019pruning, wang2019structured, fan2019reducing}, while little work attempts to perform model compression before fine-tuning.


In fact, compressing pre-trained Transformers before the fine-tuning phase is more significant.
For one thing, computational overhead of pre-trained Transformers during fine-tuning is usually heavier than that during inference, because of extra computations for gradient descent.
Furthermore, compressing pre-trained Transformers before fine-tuning can reduce the overhead during both fine-tuning and inference, which is more helpful.  

In this paper, we make the first attempt to conduct model compression for deep pre-trained Transformers before fine-tuning.
According to previous work \citep{kovaleva-etal-2019-revealing,NIPS2019_9551}, deep pre-trained Transformers are overparameterized, and only part of attention heads are actually useful for downstream tasks. 
Therefore, we propose to prune the unnecessary attention heads of pre-trained Transformers to reduce the overhead.

We train a pruner to measure the importance of attention heads and identify the unnecessary ones.
We assume the unnecessary attention heads cannot provide useful information, and pruning them will have little effect on the distribution of the text representations learned by the pre-trained Transformers.
Therefore, we design a self-supervised objective function for the pruner, which trains the model to maintain the distribution of representations after pruning.
In addition, to make our pruner more general, we adopt a meta-learning paradigm to train it. To provide diverse task distributions, we sample data from multiple corpora to form the training set of meta-learning. 



Our pruning strategy is single-shot, which means it can compress the pre-trained Transformers once before fine-tuning rather than using an iterative optimization procedure \citep{mccarley2019pruning,voita2019analyzing}.
We name our model Single-Shot Meta-Pruner (SMP).
In the experiments, we apply SMP to the representative pre-trained Transformer BERT, and conduct evaluations on GLUE \citep{wang2019glue} and the semantic relatedness tasks of SentEval \citep{conneau2018senteval}.
Experimental results show that SMP can prune as many as 50\% of attention heads of BERT without sacrificing much performance on GLUE, and even bring performance improvement on the semantic relatedness tasks of SentEval.
In addition, SMP is also comparable to, if not slightly better than, the baseline method which conducts model compression after fine-tuning.
Moreover, we find the patterns of unnecessary heads learned by SMP are transferable, which means SMP could work with different Transformer models and downstream tasks.

\section{Related Work}


To compress pre-trained Transformers,
there are two mainstream approaches, namely knowledge distillation and parameter pruning.

(1) Knowledge distillation~\cite{sanh2019distilbert,chen2020adabert,sun2020mobilebert} treats the original large model as a teacher to teach a lightweight student network. \citet{sun2019patient} design the student networks to learn from multiple intermediate layers of the teacher model. \citet{jiao2019tinybert} propose to learn from both teachers' hidden states and attention matrices. 

(2) Parameter pruning aims to remove unnecessary parts of networks, such as weight magnitude pruning~\cite{mccarley2019pruning,li2020train} and layer pruning~\cite{fan2019reducing,sajjad2020poor}. 
Given a complete BERT after fine-tuning, \citet{NIPS2019_9551} propose to prune attention heads according to the change of loss function when slightly perturbing the attention matrices. They argue that the loss function is situated in a local minimum after fine-tuning and sensitive to the change of important attention heads. Compared to this work, our SMP meta-learns the criterion and prunes PLMs before fine-tuning.

In addition to these two approaches, researchers also explore other methods, such as weight factorization~\cite{wang2019structured}, weight sharing~\cite{lan2019albert}, and parameter quantization~\cite{zafrir2019q8bert}. Most of current compression studies focus on reducing the overhead of inference.
There are also some researches trying to directly compress the models during pre-training~\citep{gordon2020compressing, sanh2019distilbert}, 
but this kind of compression has severe impact on the performance of downstream tasks. In this work, our SMP aims to reduce the overhead of both fine-tuning and inference and better maintain the performance of PLMs.

\begin{figure*}[t]
    \centering
    \includegraphics[width=\linewidth]{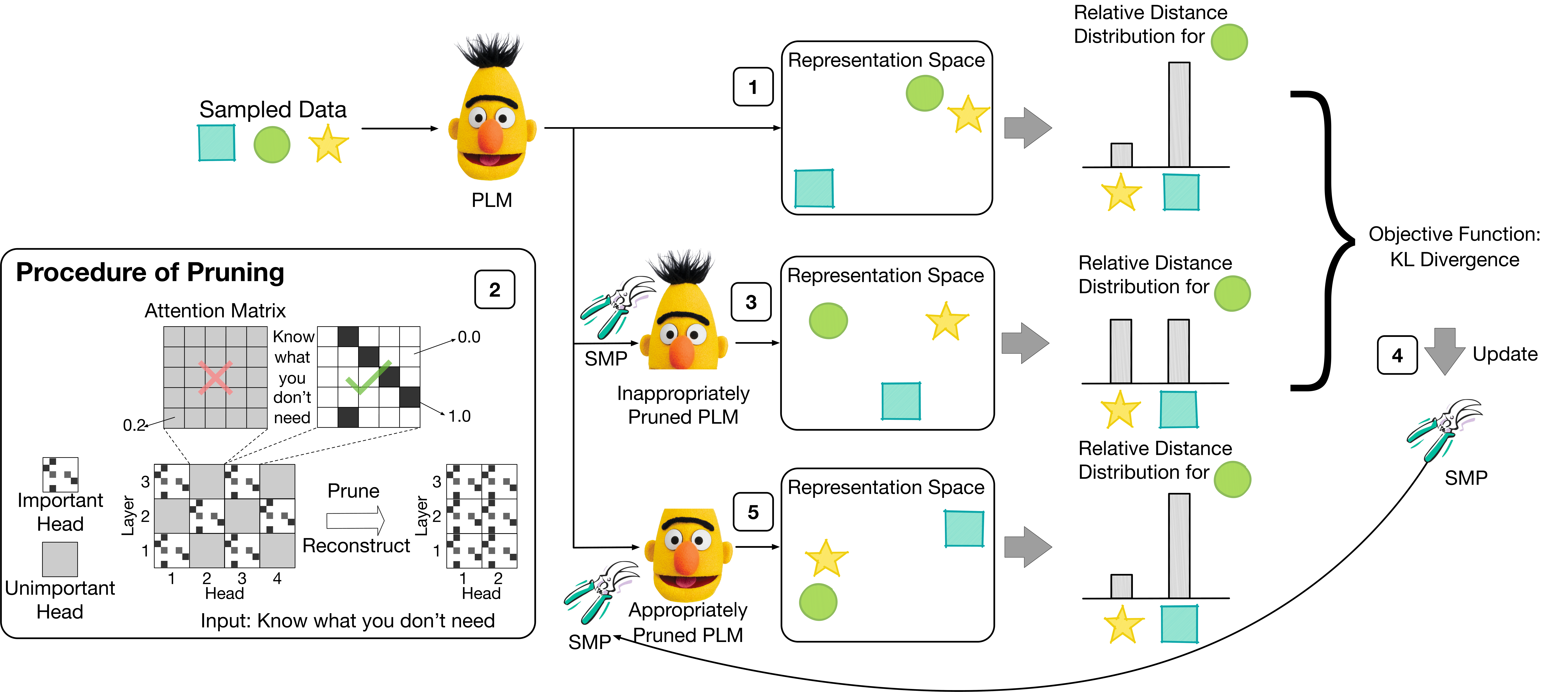}
    \captionsetup{font={normalsize}}
    \caption{An example of training our SMP and pruning a 3-layer 4-head Transformer by SMP. (1) Given the sampled data from the training corpora, a full pre-trained Transformer represents these instances in its representation space. (2) Based on the attention behaviors when encoding these instances, our SMP identifies the unnecessary heads and prunes the model. (3) In the beginning, the representation distribution of the pruned model is much different from that of the full model. (4) We use the relative distance distribution to parameterize the representation space and compute the KL divergence between distributions. (5) After optimization, the SMP can provide a good pruned model, which maintains the distribution.}
    \label{fig:framework}
\end{figure*}

From the more general perspective of pruning neural networks, our SMP prunes models before training (fine-tuning), which is different from pruning after and during training.
Pruning after training aims to identify unnecessary parts in a fully trained model based on weight magnitude~\cite{NIPS2015_5784} or effects on the loss~\cite{lecun1990optimal}. Pruning during training~\cite{louizos2018learning,voita2019analyzing} attempts to combine pruning and training procedures together. These methods require approximately the same computational overhead as training a full network.

Single-shot pruning~\cite{lee2018snip,lee2019signal,dettmers2019sparse}, which prunes networks before training, is more efficient than traditional pruning, which leads to lower computational overhead. 
Most existing studies of single-shot pruning focus on the weight pruning of randomly initialized networks by pre-defined criteria, but the models pruned by weight pruning are difficult to accelerate~\citep{NIPS2015_5784}.
In this work, we focus on directly pruning the structures (attention heads) in Transformers, which makes pruned models easy to accelerate.
We also consider how to maintain the knowledge in pre-trained models, which is different from pruning randomly initialized networks.


\section{Method}

In this section, we elaborate on our SMP model as well as its objective function and training method. Figure \ref{fig:framework} illustrates the overall framework and workflow of SMP.

The goal of SMP is to find and prune the unimportant attention heads in pre-trained Transformers.
To this end, SMP calculates the importance score of each attention head.
Then the attention heads with low importance scores are pruned to obtain a pruned Transformer.


To train SMP, we design a self-supervised objective function, which aims to keep the output of the PLMs not changing a lot after pruning.
Specifically, we propose to preserve the distribution of text representations.

Furthermore, we adopt the meta-learning paradigm in training to make SMP general and be able to apply to almost all sentence-level tasks.

\subsection{Score Calculation}
\label{sec:score}
Transformer is composed of a stack of identical layers, and each layer has two sub-layers: a multi-head self-attention network and a point-wise feed-forward network.
For the multi-head self-attention networks, each attention head yields an attention matrix.
For example, as illustrated in the bottom left of Figure \ref{fig:framework}, a 3-layer 4-head Transformer has $3\times4=12$ attention heads in total. 
To obtain the importance score of an attention head, we first compute the importance scores of its attention matrices for all instances, and then average them out as the final result.

We formulate importance score calculation of an attention matrix as an image classification task, whose input is an attention matrix and output is its importance score.
We adopt a convolutional neural network (CNN) as the encoder of attention matrices, which is widely used in image processing.
SMP concatenates a sigmoid non-linear function to the matrix encoder to output a score ranging from 0 to 1. 

Considering the difference between single-sentence and sentence-pair downstream tasks, 
SMP actually outputs a two-dimensional vector 
comprising two scores $s_{sing}$ and $s_{pair}$, which are designed for single-sentence and sentence-pair tasks respectively.
Formally, the importance score of an attention matrix is calculated by
\begin{equation}
\setlength{\abovedisplayskip}{3pt}
\setlength{\abovedisplayshortskip}{3pt}
\setlength{\belowdisplayskip}{3pt}
\setlength{\belowdisplayshortskip}{3pt}
    \begin{aligned}
        \left[s_{sing}, s_{pair}\right] &= \sigma(\text{CNN}(\mathbf {M}_{att})),
    \end{aligned}
\label{eq:score}
\end{equation}
where $\mathbf{M}_{att}$ represents an attention matrix.

After calculating the importance score of each attention head in the full pre-trained Transformer $T$, we can prune the unimportant heads and obtain a pruned Transformer $\hat T$:
\begin{equation}
\setlength{\abovedisplayskip}{5pt}
\setlength{\abovedisplayshortskip}{5pt}
\setlength{\belowdisplayskip}{3pt}
\setlength{\belowdisplayshortskip}{3pt}
    \begin{aligned}
        \hat T &= \text{SMP}(T).
    \end{aligned}
\end{equation}

\subsection{Self-supervised Objective Function}
\label{sec:obj}
Considering pre-trained Transformers essentially encode input instances into vector representations, it is reasonable to assume that the unimportant heads have little effect on the distribution of the representations of a set of sampled instances.
In other words, the representation distribution of sampled instances should be maintained after pruning those unimportant attention heads.
For example, given three instances $\{x_i, x_j, x_k\}$, we compute their representations before and after pruning by
\begin{equation*}
\setlength{\abovedisplayskip}{3pt}
\setlength{\abovedisplayshortskip}{3pt}
\setlength{\belowdisplayskip}{3pt}
\setlength{\belowdisplayshortskip}{3pt}
    \begin{aligned}
        \mathbf{h}_i &= T(x_i),\quad \mathbf{\hat h}_i &= \hat T(x_i).
    \end{aligned}
\end{equation*}
An appropriately pruned model should make sure that
if $\mathbf{h}_i$ is closer to $\mathbf{h}_j$ than $\mathbf{h}_k$, $\mathbf{\hat h}_i$ should also be closer to $\mathbf{\hat h}_j$ than $\mathbf{\hat h}_k$.

Based on this assumption, we design the training objective function for SMP.
We parameterize the representation distribution of a set of sampled instances using the \textit{relative distance distribution}.
The relative distance distribution for an instance records the normalized distances between the instance and other instances.

Given a set of instances $\{x_1, \ldots, x_N\}$, the relative distance distribution for an instance $x_n$ is an N-dimensional normalized vector $\mathbf{r}^n$, whose $i$-th entry is the relative distance between $x_n$ and $x_i$:

\begin{equation}
    \setlength{\abovedisplayskip}{3pt}
    \setlength{\abovedisplayshortskip}{3pt}
    \setlength{\belowdisplayskip}{3pt}
    \setlength{\belowdisplayshortskip}{3pt}
	\mathbf{r}^n_i = \frac{e^{\text{Dist}(\mathbf{h}_n, \mathbf{h}_i)}}{\sum_{j=1}^N e^{\text{Dist}(\mathbf{h}_n, \mathbf{h}_j)}}, 
\end{equation}
where $\text{Dist}$ is the function measuring the distance between two representations.
In this work, we simply use cosine distance.

To quantify the variation of relative distance distribution after pruning, we use the Kullback-Leibler (KL) divergence \citep{kullback1951information}.
The KL divergence between the relative distance distributions associated with the original and pruned Transformers are 
\begin{equation}
\setlength{\abovedisplayskip}{3pt}
\setlength{\abovedisplayshortskip}{3pt}
\setlength{\belowdisplayskip}{3pt}
\setlength{\belowdisplayshortskip}{3pt}
    D_{KL}(\mathbf{r}^n||\mathbf{\hat{r}}^n) = -\sum_{i=1}^N \mathbf{r}^n_i \ln(\frac{\mathbf{\hat{r}}^n_i}{\mathbf{r}^n_i}),
\end{equation}
where $\mathbf{{r}}^n$ and $\mathbf{\hat{r}}^n$ denote relative distance distributions associated with the original and pruned Transformers respectively.

Our SMP intends to maintain the representation distribution after pruning, which means making $D_{KL}(\mathbf{{r}}^n||\mathbf{\hat{r}}^n)$  as small as possible for all instances.
Therefore, the objective function of SMP is 
\begin{equation}
	 \mathcal{L}_{SMP}=\sum_{n=1}^N D_{KL}(\mathbf{{r}}^n||\mathbf{\hat{r}}^n).
\end{equation}


\subsection{Model Training via Meta-learning}
\label{sec:meta}
To train our SMP, we design a meta-learning process. We show a simple example of this training paradigm in Figure~\ref{fig:framework}. 

At the beginning of each episode, we sample $k$ instances from the training data to construct a mini dataset, which is a set of sentence pairs or single sentences. 

During pruning, we first compute the importance score for each head according to the type of the mini dataset (sentence-pair or single-sentence), as in Equation \eqref{eq:score}. 
Then we apply Gumbel-softmax~\cite{jang2016categorical} to the importance scores of all heads, which is a common reparameterization method and can transform the importance scores to discrete 0 or 1.
We multiply the outputs of an attention head by its discrete importance score of 0 or 1, and the unimportant heads, whose importance scores are 0, will be pruned.
Meanwhile, Gumbel-softmax can make sure that the pruning operation is differentiable and we can conduct back-propagation for SMP.

After pruning the pre-trained Transformer for different synthetic mini datasets, SMP is trained to adapt to different corpora and master the meta-knowledge about pruning attention heads.

\section{Experiments}


In this section, we evaluate our SMP on GLUE~\cite{wang2019glue} and SentEval~\cite{conneau2018senteval}. The pre-trained Transformers used here are  $\text{BERT}_{\text{BASE}}$ and $\text{BERT}_{\text{LARGE}}$.\footnote{https://github.com/google-research/bert}

\subsection{Experiment Setup}
\label{sec:setup}

\textbf{SMP architecture.} We set the size of input attention matrices to $(128\times 128)$, which can cover most downstream tasks. Considering some tasks, such as question answering with some input sequences longer than 128, we downsize the attention matrices. Our SMP is composed of five CNN layers. The dimension of the output of the first layer is $8$, and the following layers' dimensions are twice as large as the former layers. As a result, the dimension of the output representation for the attention matrix is $8\times 2^4=128$. To compute the matrix scores $s_{pair}$ and $s_{simp}$, we feed the output representation to a full-connected layer. 

\textbf{Pruning.} In this work, we follow the head pruning paradigm of~\cite{NIPS2019_9551} and prune the same number of heads for each layer. We set the pruning ratio to 50\%, which can significantly improve the computation efficiency and effectively maintain the original performance according to our experiments. We also report the influence of pruning ratio in this section.


\textbf{Training data for meta-learning.} We select seven datasets from GLUE~\cite{warstadt2018neural, socher2013recursive, dolan2005automatically, williams2018broad, rajpurkar2016squad, dagan2006pascal} as the training data. The statistics are shown in Table~\ref{tab:stat}. 
In particular, we split the original training data of these datasets into the training and validation part in the ratio of 9:1. 

\textbf{Training details of meta-learning.} We use two kinds of sequence-level representations of BERT: \textbf{[CLS] token} representation for sentence-pair data and \textbf{mean pooling} on the sequence outputs for single sentence data. For each training episode, we set the number of sampled instances $k$ to $60$, which lets SMP make full use of the memory of the GPUs.
The model is updated every $8$ episodes. The quicker update cycle will lead to an unstable training and the slower update cycle will bring extra training time. We choose Stochastic Gradient Descent as the optimizing algorithm and the best learning rate on the validation set is picked from $\{1,2,5\}\times 10^{-2}$. Based on the observation in the experiments, we set the total number of episodes to $48,000$, which is enough for the full convergence of SMP.
We choose the checkpoint with the lowest loss on the validation set as the final model. 
We train two SMP models based on $\text{BERT}_{\text{BASE}}$ and $\text{BERT}_{\text{LARGE}}$ respectively. SMP was trained on four 16-GB V100 GPUs for approximate 6 hours using $\text{BERT}_{\text{BASE}}$ and 18 hours using $\text{BERT}_{\text{LARGE}}$. 

\begin{table}[t]
    \centering
    \scriptsize
    \resizebox{\linewidth}{!}{
    \begin{tabular}{l|ccccccc}
    \toprule
        Dataset & MNLI & QQP & SST-2 & CoLA & STS-B & MRPC & RTE \\
    \midrule
        Type & Pair & Pair & Sing & Sing & Pair & Pair & Pair \\
    \midrule
        Size & 392k & 363k & 67k & 8.5k & 5.7k  & 3.5k  & 2.5k \\
    \bottomrule
    \end{tabular}}
    \caption{Statistics of the corpora used to train SMP. Sing refers to the single sentence task. Pair refers to the sentence-pair task.}
    \label{tab:stat}
\end{table}

\begin{table*}[t]
\centering
\scriptsize
\resizebox{\textwidth}{!}{
\begin{tabular}{ll|ccccccc|c}
  \toprule
  Model & Pruning Method & MNLI-(m/mm) & QQP & SST-2 & CoLA & STS-B & MRPC & RTE & Average \\
\midrule
 & None & 83.85/83.82 & 90.96 & 92.43 & 57.84 & 88.71 & 85.78 & 65.70 & \underline{81.14} \\
  \cmidrule{2-10}
  & $L_0$ Norm & 79.70/79.83 & 85.82 & 91.74 & 52.10 & 88.30 & 77.45 & 62.45 & 77.17 \\
  & HISP & 81.69/81.90 & 86.88 &  91.85 & 54.84 & 88.46 & 81.12 & 65.34 & 79.01\\
  $\text{BERT}_{\text{BASE}}$ & HISP-retrain & 83.56/83.73 & 91.03 & 92.20 & 53.24 & 88.58 & 85.04 & 66.78 & 80.52 \\
  \cmidrule{2-10}
  & Random & 82.43/82.63 & 90.34 &  91.83 & 52.37 & 87.83 & 80.88 & 65.77 & 79.26 \\
  & SMP & 83.36/83.75 & 90.96 & 92.31 & 57.26 & 88.49 & 85.04 & 67.87 & \textbf{81.13} \\
  \midrule
 & None & 87.87/87.62 & 91.49 & 93.69 & 63.89 & 90.99 & 88.72 & 85.92 & \underline{86.27} \\
  \cmidrule{2-10}
  & $L_0$ Norm & 85.93/85.83 & 90.26 & 93.46 & 56.02 & 90.33 & 86.51 & 81.94 & 83.79\\
  & HISP & 85.44/85.67 & 85.17 & 93.11 & 62.54 & 89.65 & 87.50 & 81.58 & 83.83 \\
  $\text{BERT}_{\text{LARGE}}$ & HISP-retrain & 87.42/87.26 & 91.55 & 93.46 & 60.09 & 90.14 & 89.70 & 83.03 & 85.33\\
  \cmidrule{2-10}
  & Random & 86.36/86.43 & 91.26 &  92.45 & 58.78 & 90.31 & 86.85 & 80.35 & 84.10 \\
  & SMP & 86.86/86.96 & 91.46 & 93.34 & 63.57 & 90.95 & 89.70 & 82.31 & \textbf{85.64} \\
  \bottomrule
\end{tabular}
}
\caption{Results on seven tasks in GLUE (\%). HISP-retrain is an inference-oriented pruning method, which only reduces the overhead of inference. SMP is our method, which reduces the overhead of both fine-tuning and inference. MNLI contains two validation sets and provides two results. We underline the truly best results (from the original models) and boldface the best results among the pruned models.}
\label{tab-glue}
\end{table*}


\textbf{Baselines.} To validate the effectiveness of SMP, we introduce four baselines in our experiments.

(1) \textbf{Fine-tune (None).} We fine-tune a complete BERT on downstream tasks, which can provide an oracle result without pruning. 

(2) \textbf{Random.} We randomly select the same number of heads to prune as SMP in each layer before fine-tuning. We repeat the random experiments five times and report the mean of model performances. Since the number of head combinations is very large, random experiments only give a rough estimation of performances. For example, $\text{BERT}_{\text{BASE}}$ has 144 attention heads, and there are $C_{144}^{72} \ge 10^{42}$ combinations for the pruning ratio of 50\%. 

(3) \textbf{$L_0$ Norm.} Following~\citet{voita2019analyzing}, we multiply the output of each head with a scalar gate and introduce an $L_0$ regularization loss to these gates. Using this method, we can search the optimal value of each gate by gradient descent.

(4) \textbf{HISP and HISP-retrain.} 
Besides $L_0$ Norm, we adopt the attention head pruning method introduced by~\citet{NIPS2019_9551}, called Head Importance Score for Pruning (HISP). 
The original algorithm directly evaluates the model performance after pruning. According to previous studies on general neural network pruning~\cite{NIPS2015_5784}, retraining after pruning can further promote the performance of pruned models. Hence, we introduce {HISP-retrain}, which retrains pruned models for better performance. In our experiments, we retrain the pruned model given by HISP for $3$ additional epochs as HISP-retrain.
Since HISP prunes models after fine-tuning, pre-trained Transformers could better learn from fine-tuning due to the larger model capacity.

\subsection{GLUE}

The GLUE benchmark~\cite{wang2019glue} is used to validate the effectiveness of SMP on the general fine-tuning tasks. We compare four methods mentioned above on seven downstream tasks in the GLUE benchmark. We exclude two tasks in GLUE, namely the Winograd Schema Challenge and QNLI. The former is excluded due to the small size of the dataset while the latter is excluded for the experiment on model transferability. The fine-tuning experiments follow the hyperparameters reported in the original study~\cite{devlin2018bert} except the number of epochs. The random baseline and SMP adopt the same hyper-parameters used in fine-tuning a complete BERT. For small datasets containing less than 10,000 instances, we set the number of epochs to 10. For the others, we keep the original number unchanged (3 epochs).

We report the results on the validation, rather than test data, so the results differ from the original BERT paper.
From Table~\ref{tab-glue}, we observe that: 

(1) The average performance of random pruning is consistently worse than that of fine-tuning, which shows the serious impact of pruning on pre-trained Transformers. In the experiments, we find that some random seeds lead to a good performance while some random seeds significantly degrade model performance. 
The variation of random pruning proves the assumption that there are important attention heads, which should not be pruned before fine-tuning, in pre-trained Transformers. It is related to the lottery ticket hypothesis for pre-trained Transformers~\cite{chen2020lottery}.

\begin{table*}[t]
    \centering
    \small
    \begin{tabular}{ll|ccccccc}
    \toprule
    Model  &  Pruning Method & STS-12 & STS-13 &  STS-14 & STS-15 & STS-16 & STS-B & SICK-R  \\
    \midrule
    GloVe BoW & --- & 52.10 & 49.60 & 54.60 & 56.10 & 51.40 &  64.70 & 79.90 \\
   InferSent & --- & 59.20 & 58.90 & 69.60 & 71.30 & 71.50 &  \underline{75.60} & \underline{88.30} \\
    \midrule
      \multirow{3}{*}{$\text{BERT}_{\text{BASE}}$} & Full & 46.87 & 52.77 & 57.15 & 63.46 & 64.50 & 65.49 & 80.57 \\
    \cmidrule{2-9}
     & Random & 51.07 & 48.19 & 57.66 & 64.48 & 61.00 & 65.98 & 79.67\\
         & SMP & \textbf{57.59} & \underline{\textbf{63.94}} & \textbf{64.64} & \textbf{69.06} & \textbf{66.80} & \textbf{70.18} & \textbf{82.19} \\
    \midrule
      \multirow{3}{*}{$\text{BERT}_{\text{LARGE}}$} & Full & 54.87 & 60.78 & 64.21 & 68.07 & 66.65 & 69.91 & 83.91 \\
    \cmidrule{2-9}
     & Random & 55.47 & 55.04 & 63.85 & 67.70 & 64.47 & 70.42 & 83.18\\
         & SMP & \underline{\textbf{62.13}} & \textbf{62.57} & \underline{\textbf{71.18}} & \underline{\textbf{74.38}} & \underline{\textbf{71.55}} & \textbf{71.19} & \textbf{84.52} \\
    \bottomrule
    \end{tabular}
    \caption{Results on semantic relatedness tasks in SentEval. Numbers reported are Pearson correlations x100. The results of GloVe and InferSent are from the paper of SentEval~\cite{conneau2018senteval}. We underline the overall best results and boldface the best results among BERT models.}
    \label{tab:senteval}
\end{table*}

(2) The overall performances of $L_0$ Norm and HISP are worse than that of random pruning, which indicates that head pruning on a converged model leads to serious performance degradation. 
Meanwhile, we find that HISP-retrain significantly outperforms HISP, which reflects the importance of retraining in pruning-after-training approaches. 
Most results of HISP-retrain are better than those of random pruning and close to the result of fine-tuning, which indicates that HISP-retrain can select important heads for downstream tasks and provide a good pruned model. However, there still exists the case that retraining degrades the performance. For $\text{BERT}_{\text{BASE}}$, the performance of  HISP-retrain on CoLA is lower than HISP by about $1.5\%$.

(3) SMP achieves the best results on average among these pruning methods, and have comparable performance with fine-tuning, which indicates SMP significantly reduces the impact of pruning on downstream tasks. Besides, SMP even outperforms the fine-tuning method in some tasks, such as RTE for $\text{BERT}_{\text{BASE}}$ and MRPC for $\text{BERT}_{\text{LARGE}}$. It shows that \textbf{pruning unnecessary structure can also promote the performance of downstream tasks of Transformers.} Besides, SMP works well on both $\text{BERT}_{\text{BASE}}$ and $\text{BERT}_{\text{LARGE}}$, which reveals the generality of SMP. 

\begin{table}[t]
    \centering
    \small
    \resizebox{0.48\textwidth}{!}{
    \begin{tabular}{lr|rr|rr}
    \toprule
       Model & Ratio & \multicolumn{2}{c|}{Memory (MB)} & \multicolumn{2}{c}{Speed (IPS)}  \\
    \midrule
    \multirow{2}{*}{$\text{BERT}_{\text{BASE}}$}  &$0\%$ & 841 & --- & 18.2 & --- \\
        &$50\%$ & 538 & $-36.0\%$ & 24.5 & $+34.6\%$ \\
    \midrule
    \multirow{2}{*}{$\text{BERT}_{\text{LARGE}}$}  &$0\%$ & 2,156 & --- & 5.3 & --- \\
        &$50\%$ & 1,514 & $-29.8\%$ & 7.3 & $+37.7\%$\\
    \bottomrule
    \end{tabular}
    }
    \caption{Average memory overhead per instance and the speed in instances per second (IPS) on QNLI.}
    \label{tab:speed}
\end{table}

\begin{table*}[t]
\centering
\scriptsize
\resizebox{.9\textwidth}{!}{
\begin{tabular}{l|ccccccc|c}
  \toprule
  SMP Model & MNLI-(m/mm) & QQP & SST-2 & CoLA & STS-B & MRPC & RTE & Average \\
\midrule
  SMP-LARGE & 86.86/86.96 & 91.46 & 93.34 & 63.57 & 90.95 & 89.70 & 82.31 & 85.64 \\
  SMP-BASE & 86.72/86.63 & 91.43 & 93.23 & 64.79 & 91.00 & 89.46 & 83.39 & \textbf{85.83} \\
  \bottomrule
\end{tabular}
}
\caption{Results of transferability on seven tasks in GLUE (\%). We prune $\text{BERT}_{\text{LARGE}}$ using two different SMP models. SMP-LARGE is the model trained on $\text{BERT}_{\text{LARGE}}$ while SMP-BASE is the model trained on $\text{BERT}_{\text{BASE}}$.}
\label{tab-trans-model}
\end{table*}

\subsection{SentEval}

SentEval~\cite{conneau2018senteval} is used to validate the representation ability of the models pruned by SMP. The goal of SMP is to preserve the distribution of text representations after pruning for maintaining important prior knowledge learned by pre-training. Hence, we use SentEval to investigate whether SMP maintains important prior knowledge after pruning. As mentioned in the experimental setup, there are two approaches to compute text representations from pre-trained BERT. Based on the findings of previous work~\cite{ma2019universal}, \textbf{mean pooling} is better than \textbf{[CLS] token} so that we use the representations of \textbf{mean pooling} in SentEval experiments. We use the sentence relatedness tasks in SentEval to evaluate the unsupervised representation ability of pruned models because these tasks directly use text representations to compute the cosine similarity without additional architecture and training. Sentence relatedness tasks are composed of six STS tasks~\cite{agirre2012semeval,agirre2013sem,agirre2014semeval,agirre2015semeval,cer2017semeval} and SICK-R~\cite{marelli2014semeval}. Since these tasks are unsupervised, we exclude $L_0$ Norm and HISP, which need supervision.


We report the results on semantic relatedness tasks in Table~\ref{tab:senteval}. From this table, we have two observations: (1) Random pruning has similar performance to a full BERT. In some tasks, randomly pruned models are even better than full models. It indicates a full BERT cannot provide good text representations for semantic relatedness although there is rich language knowledge in the pre-trained model. For these tasks, BERT might have many unnecessary heads so even random pruning could bring in minor performance improvements. (2) SMP significantly improves the performance of the pruned BERT, which indicates \textbf{SMP makes full use of important prior knowledge} and helps pruned models provide informative representations for unsupervised tasks. 

\subsection{Effect on Fine-tuning Efficiency}

In this subsection, we investigate the effect of pruning on fine-tuning efficiency, which is the advantage of single-shot pruning.
Experiments are conducted on a machine equipped with Tesla P40 GPUs.
As shown in Table~\ref{tab:speed}, pruning 50\% of the model's heads speeds up fine-tuning by more than $34\%$ and reduce the memory overhead per instance by around $30\%$. In this case, we train more instances simultaneous for the pruned model due to the less memory overhead. Besides, the time of running SMP is nearly 1/30 of the fine-tuning time, which is negligible. According to the results, single-shot pruning on deep Transformers can make fine-tuning more efficient.

\begin{figure}
    \centering
    \includegraphics[width=0.8\linewidth]{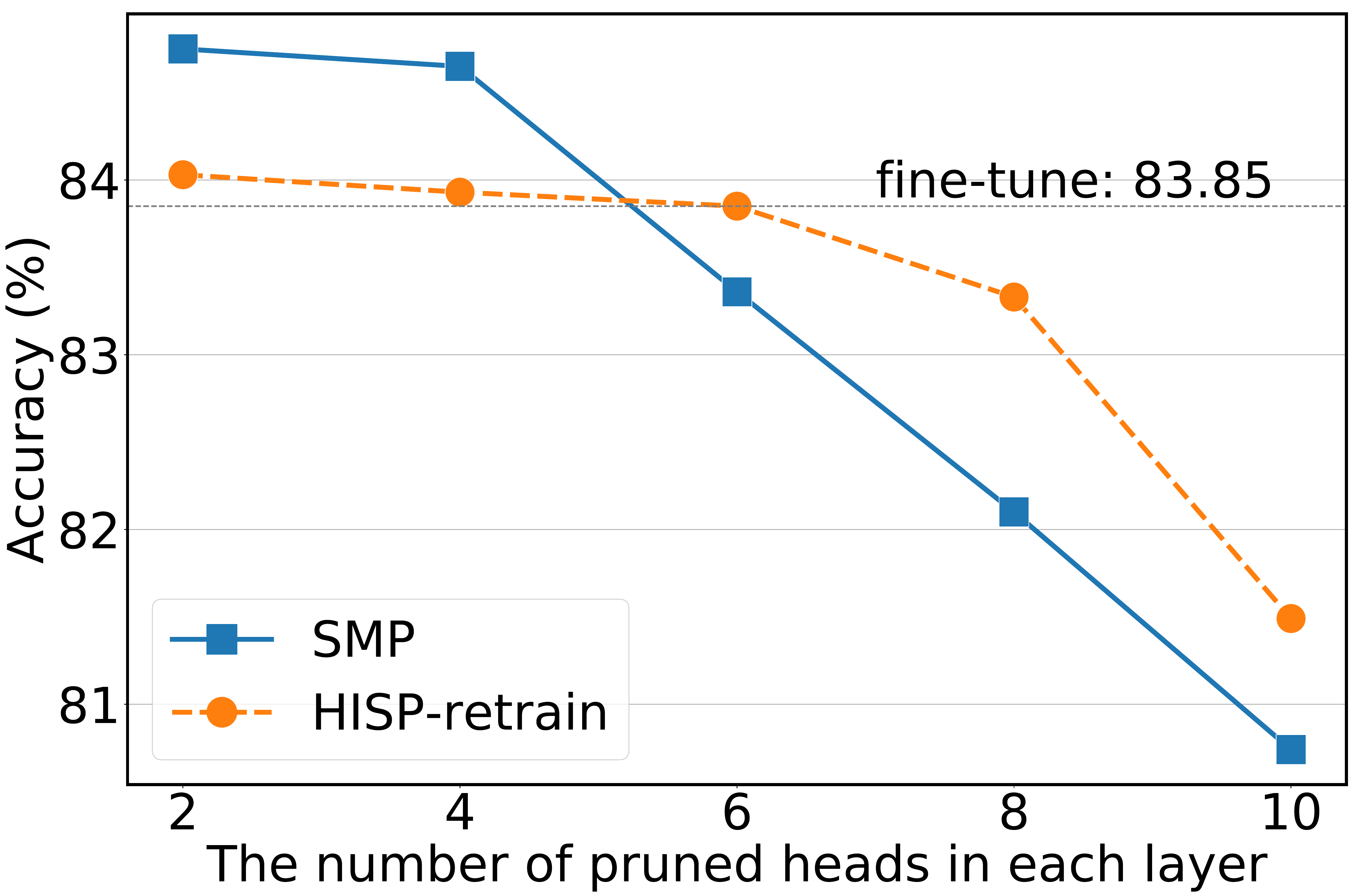}
    \caption{Evolution of accuracy on the MultiNLI-matched when heads are pruned from $\text{BERT}_{\text{BASE}}$.}
    \label{fig:pruned-ratio}
\end{figure}

\begin{figure*}[t]
    \centering
    \includegraphics[width=\linewidth]{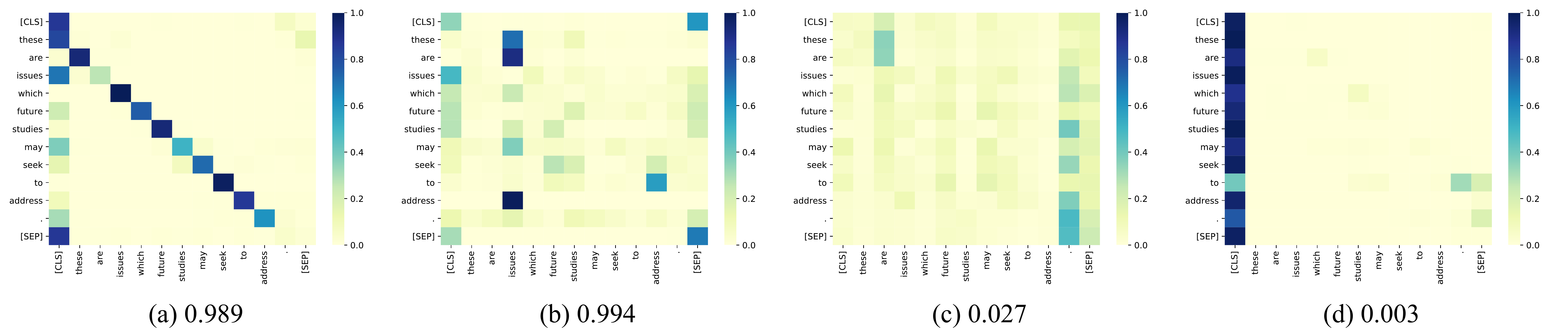}
    \caption{Detection of the implicit pruning rules learned by SMP. Given input sentence ``These are issues which further studies may seek to address.'', we present four attention matrices with their corresponding scores as subtitles. The left two matrices' scores are close to 1 while the right two matrices' scores are close to 0.}
    \label{fig:matrices}
\vspace{-0.2em}
\end{figure*}

\subsection{Influence of Pruning Ratio}
\label{sec:ratio}

In this subsection, we investigate the influence of pruning ratio. We test pruned models on the MultiNLI-matched validation set. As shown in Figure~\ref{fig:pruned-ratio}, we observe that: (1) Pruning a small number of unnecessary attention heads promotes the performance by nearly 1\%. (2) SMP is better when the number of pruned heads is smaller than 6 while HISP-retrain is better in the other cases. 
It indicates that pruning too many parameters before fine-tuning will influence the performance on downstream tasks where the process of retraining could save this degradation of performance.

\subsection{Transferability}
\label{sec:trans}

In this part, we evaluate the transferability of SMP. We consider two kinds of transferability, including the transferability to new Transformer encoders and new datasets.

\textbf{Transferability to New Transformer Encoders.} We use the SMP trained on $\text{BERT}_{\text{BASE}}$ to prune $\text{BERT}_{\text{LARGE}}$. The results are shown in Table~\ref{tab-trans-model}. We observe that the SMP trained on $\text{BERT}_{\text{BASE}}$ achieves comparable results to the SMP trained on $\text{BERT}_{\text{LARGE}}$ when pruning $\text{BERT}_{\text{LARGE}}$. It indicates that the attention patterns learned by SMP are general in the Transformer encoders with different sizes.

\textbf{Transferability to New Datasets.} We choose QNLI, which is a natural language inference (NLI) dataset, as the task. QNLI is used to validate whether SMP can transfer to a new NLI dataset. Note that we use $\text{BERT}_{\text{BASE}}$ as the pre-trained Transformer. As shown in Table~\ref{tab:trans}, SMP increases the average performance of random pruning and achieves comparable result to HISP-retrain. It demonstrates that SMP captures general patterns in attention matrices, which can transfer to pruning pre-trained Transformers on other tasks.

\begin{table}[t]
    \centering
    \scriptsize
    \resizebox{\linewidth}{!}{
    \begin{tabular}{l|cc|cc}
    \toprule
        Method & Fine-tune & HISP-retrain & Random & SMP \\
    \midrule
    Acc. & 91.3 & 91.0 & 89.4 & 90.6 \\
    \bottomrule
    \end{tabular}}
    \caption{Accuracy on QNLI.}
    \label{tab:trans}
\vspace{-0.2em}
\end{table}





\subsection{Visualization}

In this subsection, we investigate the implicit rules learned by SMP. We compute the attention matrices for a given sentence, and score each attention matrix. In Figure~\ref{fig:matrices}, we show four attention matrices. The first matrix seems to be a lower diagonal matrix, which refers to the attention to previous words. This attention head implicitly captures the sequential information of the sentence. The second matrix shows a strong relation between ``address'' and ``issues'', which is long-term dependency. In the third matrix, every element is small. Therefore, this matrix does not bring any useful information. In the last matrix, although there are extremely high values, all tokens attend to the same token ([CLS]), which is not informative. SMP gives high scores to the first two matrices and low scores to the last two matrices, which shows the implicit pruning rules learned by SMP are consistent with human intuition.

\section{Conclusion and Future Work}

In this work, we propose Single-Shot Meta-Pruning to reduce the computational overhead of both fine-tuning and inference when using deep pre-trained Transformers. Specifically, SMP learns the implicit rules for pruning in terms of attention matrices and adaptively prunes unnecessary attention heads before fine-tuning. In our experiments, we find pruning 50\% of attention heads with SMP has little impact on the performances on downstream tasks. What's more, pruning a few unnecessary heads can further improve the model performance in some cases.

There are four important directions for future research: (1) Explore task-aware pruning, such as taking the labels of instances into account. 
(2) Joint pruning in each layer to maintain more diversity in pruned models, such as limiting the number of the attention heads sharing similar patterns in each layer. (3) Discover more unnecessary structures in Transformer, such as point-wise feed-forward networks. (4) Apply implicit pruning rules to constraining the pre-training procedure of Transformers, which guides pre-training models through the more efficient use of parameters and attention.

\bibliography{emnlp2020}

\begin{thebibliography}{47}
\expandafter\ifx\csname natexlab\endcsname\relax\def\natexlab#1{#1}\fi

\bibitem[{Agirre et~al.(2015)Agirre, Banea, Cardie, Cer, Diab, Gonzalez-Agirre,
  Guo, Lopez-Gazpio, Maritxalar, Mihalcea et~al.}]{agirre2015semeval}
Eneko Agirre, Carmen Banea, Claire Cardie, Daniel Cer, Mona Diab, Aitor
  Gonzalez-Agirre, Weiwei Guo, Inigo Lopez-Gazpio, Montse Maritxalar, Rada
  Mihalcea, et~al. 2015.
\newblock Semeval-2015 task 2: Semantic textual similarity, english, spanish
  and pilot on interpretability.
\newblock In \emph{Proceedings of SemEval}.

\bibitem[{Agirre et~al.(2014)Agirre, Banea, Cardie, Cer, Diab, Gonzalez-Agirre,
  Guo, Mihalcea, Rigau, and Wiebe}]{agirre2014semeval}
Eneko Agirre, Carmen Banea, Claire Cardie, Daniel Cer, Mona Diab, Aitor
  Gonzalez-Agirre, Weiwei Guo, Rada Mihalcea, German Rigau, and Janyce Wiebe.
  2014.
\newblock Semeval-2014 task 10: Multilingual semantic textual similarity.
\newblock In \emph{Proceedings of SemEval}.

\bibitem[{Agirre et~al.(2013)Agirre, Cer, Diab, Gonzalez-Agirre, and
  Guo}]{agirre2013sem}
Eneko Agirre, Daniel Cer, Mona Diab, Aitor Gonzalez-Agirre, and Weiwei Guo.
  2013.
\newblock Semeval-2013 shared task: Semantic textual similarity.
\newblock In \emph{Proceedings of SemEval}.

\bibitem[{Agirre et~al.(2012)Agirre, Diab, Cer, and
  Gonzalez-Agirre}]{agirre2012semeval}
Eneko Agirre, Mona Diab, Daniel Cer, and Aitor Gonzalez-Agirre. 2012.
\newblock Semeval-2012 task 6: A pilot on semantic textual similarity.
\newblock In \emph{Proceedings of SemEval}.

\bibitem[{Brown et~al.(2020)Brown, Mann, Ryder, Subbiah, Kaplan, Dhariwal,
  Neelakantan, Shyam, Sastry, Askell, Agarwal, Herbert-Voss, Krueger, Henighan,
  Child, Ramesh, Ziegler, Wu, Winter, Hesse, Chen, Sigler, Litwin, Gray, Chess,
  Clark, Berner, McCandlish, Radford, Sutskever, and
  Amodei}]{brown2020language}
Tom~B. Brown, Benjamin Mann, Nick Ryder, Melanie Subbiah, Jared Kaplan,
  Prafulla Dhariwal, Arvind Neelakantan, Pranav Shyam, Girish Sastry, Amanda
  Askell, Sandhini Agarwal, Ariel Herbert-Voss, Gretchen Krueger, Tom Henighan,
  Rewon Child, Aditya Ramesh, Daniel~M. Ziegler, Jeffrey Wu, Clemens Winter,
  Christopher Hesse, Mark Chen, Eric Sigler, Mateusz Litwin, Scott Gray,
  Benjamin Chess, Jack Clark, Christopher Berner, Sam McCandlish, Alec Radford,
  Ilya Sutskever, and Dario Amodei. 2020.
\newblock Language models are few-shot learners.
\newblock \emph{arXiv preprint arXiv:2005.14165}.

\bibitem[{Cer et~al.(2017)Cer, Diab, Agirre, Lopez-Gazpio, and
  Specia}]{cer2017semeval}
Daniel Cer, Mona Diab, Eneko Agirre, Inigo Lopez-Gazpio, and Lucia Specia.
  2017.
\newblock Semeval-2017 task 1: Semantic textual similarity-multilingual and
  cross-lingual focused evaluation.
\newblock \emph{arXiv preprint arXiv:1708.00055}.

\bibitem[{Chen et~al.(2020{\natexlab{a}})Chen, Li, Qiu, Wang, Li, Ding, Deng,
  Huang, Lin, and Zhou}]{chen2020adabert}
Daoyuan Chen, Yaliang Li, Minghui Qiu, Zhen Wang, Bofang Li, Bolin Ding, Hongbo
  Deng, Jun Huang, Wei Lin, and Jingren Zhou. 2020{\natexlab{a}}.
\newblock {AdaBERT}: Task-adaptive {BERT} compression with differentiable
  neural architecture search.
\newblock \emph{arXiv preprint arXiv:2001.04246}.

\bibitem[{Chen et~al.(2020{\natexlab{b}})Chen, Frankle, Chang, Liu, Zhang,
  Wang, and Carbin}]{chen2020lottery}
Tianlong Chen, Jonathan Frankle, Shiyu Chang, Sijia Liu, Yang Zhang, Zhangyang
  Wang, and Michael Carbin. 2020{\natexlab{b}}.
\newblock The lottery ticket hypothesis for pre-trained bert networks.
\newblock \emph{arXiv preprint arXiv:2007.12223}.

\bibitem[{Conneau and Kiela(2018)}]{conneau2018senteval}
Alexis Conneau and Douwe Kiela. 2018.
\newblock Senteval: An evaluation toolkit for universal sentence
  representations.
\newblock In \emph{Proceedings of LREC}.

\bibitem[{Dagan et~al.(2006)Dagan, Glickman, and Magnini}]{dagan2006pascal}
Ido Dagan, Oren Glickman, and Bernardo Magnini. 2006.
\newblock The {PASCAL} recognising textual entailment challenge.
\newblock In \emph{Machine learning challenges.}

\bibitem[{Dettmers and Zettlemoyer(2019)}]{dettmers2019sparse}
Tim Dettmers and Luke Zettlemoyer. 2019.
\newblock Sparse networks from scratch: Faster training without losing
  performance.
\newblock \emph{arXiv preprint arXiv:1907.04840}.

\bibitem[{Devlin et~al.(2019)Devlin, Chang, Lee, and
  Toutanova}]{devlin2018bert}
Jacob Devlin, Ming-Wei Chang, Kenton Lee, and Kristina Toutanova. 2019.
\newblock {BERT}: Pre-training of deep bidirectional transformers for language
  understanding.
\newblock In \emph{Proceedings of NAACL-HLT}.

\bibitem[{Dolan and Brockett(2005)}]{dolan2005automatically}
William~B Dolan and Chris Brockett. 2005.
\newblock Automatically constructing a corpus of sentential paraphrases.
\newblock In \emph{Proceedings of IWP}.

\bibitem[{Fan et~al.(2020)Fan, Grave, and Joulin}]{fan2019reducing}
Angela Fan, Edouard Grave, and Armand Joulin. 2020.
\newblock Reducing transformer depth on demand with structured dropout.
\newblock In \emph{Proceedings of ICLR}.

\bibitem[{Gordon et~al.(2020)Gordon, Duh, and Andrews}]{gordon2020compressing}
Mitchell~A Gordon, Kevin Duh, and Nicholas Andrews. 2020.
\newblock Compressing {BERT}: Studying the effects of weight pruning on
  transfer learning.
\newblock \emph{arXiv preprint arXiv:2002.08307}.

\bibitem[{Han et~al.(2015)Han, Pool, Tran, and Dally}]{NIPS2015_5784}
Song Han, Jeff Pool, John Tran, and William Dally. 2015.
\newblock Learning both weights and connections for efficient neural network.
\newblock In \emph{Proceedings of NIPS}.

\bibitem[{Jang et~al.(2016)Jang, Gu, and Poole}]{jang2016categorical}
Eric Jang, Shixiang Gu, and Ben Poole. 2016.
\newblock Categorical reparameterization with gumbel-softmax.
\newblock \emph{arXiv preprint arXiv:1611.01144}.

\bibitem[{Jiao et~al.(2019)Jiao, Yin, Shang, Jiang, Chen, Li, Wang, and
  Liu}]{jiao2019tinybert}
Xiaoqi Jiao, Yichun Yin, Lifeng Shang, Xin Jiang, Xiao Chen, Linlin Li, Fang
  Wang, and Qun Liu. 2019.
\newblock Tiny{BERT}: Distilling bert for natural language understanding.
\newblock \emph{arXiv preprint arXiv:1909.10351}.

\bibitem[{Kovaleva et~al.(2019)Kovaleva, Romanov, Rogers, and
  Rumshisky}]{kovaleva-etal-2019-revealing}
Olga Kovaleva, Alexey Romanov, Anna Rogers, and Anna Rumshisky. 2019.
\newblock Revealing the dark secrets of {BERT}.
\newblock In \emph{Proceedings of EMNLP-IJCNLP}.

\bibitem[{Kullback and Leibler(1951)}]{kullback1951information}
Solomon Kullback and Richard~A Leibler. 1951.
\newblock On information and sufficiency.
\newblock \emph{The annals of mathematical statistics}.

\bibitem[{Lan et~al.(2020)Lan, Chen, Goodman, Gimpel, Sharma, and
  Soricut}]{lan2019albert}
Zhenzhong Lan, Mingda Chen, Sebastian Goodman, Kevin Gimpel, Piyush Sharma, and
  Radu Soricut. 2020.
\newblock {ALBERT}: A lite bert for self-supervised learning of language
  representations.
\newblock In \emph{Proceedings of ICLR}.

\bibitem[{LeCun et~al.(1990)LeCun, Denker, and Solla}]{lecun1990optimal}
Yann LeCun, John~S Denker, and Sara~A Solla. 1990.
\newblock Optimal brain damage.
\newblock In \emph{Proceeding of NIPS}.

\bibitem[{Lee et~al.(2019{\natexlab{a}})Lee, Ajanthan, Gould, and
  Torr}]{lee2019signal}
Namhoon Lee, Thalaiyasingam Ajanthan, Stephen Gould, and Philip~HS Torr.
  2019{\natexlab{a}}.
\newblock A signal propagation perspective for pruning neural networks at
  initialization.
\newblock \emph{arXiv preprint arXiv:1906.06307}.

\bibitem[{Lee et~al.(2019{\natexlab{b}})Lee, Ajanthan, and Torr}]{lee2018snip}
Namhoon Lee, Thalaiyasingam Ajanthan, and Philip Torr. 2019{\natexlab{b}}.
\newblock {SNIP}: Single-shot network pruning based on connection sensitivity.
\newblock In \emph{Proceedings of ICLR}.

\bibitem[{Li et~al.(2020)Li, Wallace, Shen, Lin, Keutzer, Klein, and
  Gonzalez}]{li2020train}
Zhuohan Li, Eric Wallace, Sheng Shen, Kevin Lin, Kurt Keutzer, Dan Klein, and
  Joseph~E Gonzalez. 2020.
\newblock Train large, then compress: Rethinking model size for efficient
  training and inference of transformers.
\newblock \emph{arXiv preprint arXiv:2002.11794}.

\bibitem[{Liu et~al.(2019)Liu, Ott, Goyal, Du, Joshi, Chen, Levy, Lewis,
  Zettlemoyer, and Stoyanov}]{liu2019roberta}
Yinhan Liu, Myle Ott, Naman Goyal, Jingfei Du, Mandar Joshi, Danqi Chen, Omer
  Levy, Mike Lewis, Luke Zettlemoyer, and Veselin Stoyanov. 2019.
\newblock {RoBERTa}: A robustly optimized {BERT} pretraining approach.
\newblock \emph{arXiv preprint arXiv:1907.11692}.

\bibitem[{Louizos et~al.(2018)Louizos, Welling, and
  Kingma}]{louizos2018learning}
Christos Louizos, Max Welling, and Diederik~P. Kingma. 2018.
\newblock Learning sparse neural networks through {L0} regularization.
\newblock In \emph{Proceedings of ICLR}.

\bibitem[{Ma et~al.(2019)Ma, Xu, Wang, Nallapati, and Xiang}]{ma2019universal}
Xiaofei Ma, Peng Xu, Zhiguo Wang, Ramesh Nallapati, and Bing Xiang. 2019.
\newblock Universal text representation from bert: An empirical study.
\newblock \emph{arXiv preprint arXiv:1910.07973}.

\bibitem[{Marelli et~al.(2014)Marelli, Bentivogli, Baroni, Bernardi, Menini,
  and Zamparelli}]{marelli2014semeval}
Marco Marelli, Luisa Bentivogli, Marco Baroni, Raffaella Bernardi, Stefano
  Menini, and Roberto Zamparelli. 2014.
\newblock Semeval-2014 task 1: Evaluation of compositional distributional
  semantic models on full sentences through semantic relatedness and textual
  entailment.
\newblock In \emph{Proceedings of SemEval}.

\bibitem[{McCarley(2019)}]{mccarley2019pruning}
JS~McCarley. 2019.
\newblock Pruning a {BERT}-based question answering model.
\newblock \emph{arXiv preprint arXiv:1910.06360}.

\bibitem[{Michel et~al.(2019)Michel, Levy, and Neubig}]{NIPS2019_9551}
Paul Michel, Omer Levy, and Graham Neubig. 2019.
\newblock Are sixteen heads really better than one?
\newblock In \emph{Proceedings of NIPS}.

\bibitem[{Radford et~al.(2018)Radford, Narasimhan, Salimans, and
  Sutskever}]{radford2018improving}
Alec Radford, Karthik Narasimhan, Tim Salimans, and Ilya Sutskever. 2018.
\newblock Improving language understanding by generative pre-training.
\newblock In \emph{Proceedings of OpenAI Technical report}.

\bibitem[{Rajpurkar et~al.(2016)Rajpurkar, Zhang, Lopyrev, and
  Liang}]{rajpurkar2016squad}
Pranav Rajpurkar, Jian Zhang, Konstantin Lopyrev, and Percy Liang. 2016.
\newblock {SQ}u{AD}: 100,000+ questions for machine comprehension of text.
\newblock In \emph{Proceedings of EMNLP}.

\bibitem[{Sajjad et~al.(2020)Sajjad, Dalvi, Durrani, and
  Nakov}]{sajjad2020poor}
Hassan Sajjad, Fahim Dalvi, Nadir Durrani, and Preslav Nakov. 2020.
\newblock Poor man's {BERT}: Smaller and faster transformer models.
\newblock In \emph{Proceedings of ACL}.

\bibitem[{Sanh et~al.(2019)Sanh, Debut, Chaumond, and
  Wolf}]{sanh2019distilbert}
Victor Sanh, Lysandre Debut, Julien Chaumond, and Thomas Wolf. 2019.
\newblock Distil{BERT}, a distilled version of {BERT}: smaller, faster, cheaper
  and lighter.
\newblock \emph{arXiv preprint arXiv:1910.01108}.

\bibitem[{Socher et~al.(2013)Socher, Perelygin, Wu, Chuang, Manning, Ng, and
  Potts}]{socher2013recursive}
Richard Socher, Alex Perelygin, Jean Wu, Jason Chuang, Christopher~D Manning,
  Andrew Ng, and Christopher Potts. 2013.
\newblock Recursive deep models for semantic compositionality over a sentiment
  treebank.
\newblock In \emph{Proceedings of EMNLP}.

\bibitem[{Sun et~al.(2019)Sun, Cheng, Gan, and Liu}]{sun2019patient}
Siqi Sun, Yu~Cheng, Zhe Gan, and Jingjing Liu. 2019.
\newblock Patient knowledge distillation for bert model compression.
\newblock In \emph{Proceedings of EMNLP}.

\bibitem[{Sun et~al.(2020)Sun, Yu, Song, Liu, Yang, and
  Zhou}]{sun2020mobilebert}
Zhiqing Sun, Hongkun Yu, Xiaodan Song, Renjie Liu, Yiming Yang, and Denny Zhou.
  2020.
\newblock Mobile{BERT}: a compact task-agnostic bert for resource-limited
  devices.
\newblock In \emph{Proceedings of ACL}.

\bibitem[{Tang et~al.(2019)Tang, Lu, Liu, Mou, Vechtomova, and
  Lin}]{tang2019distilling}
Raphael Tang, Yao Lu, Linqing Liu, Lili Mou, Olga Vechtomova, and Jimmy Lin.
  2019.
\newblock Distilling task-specific knowledge from {BERT} into simple neural
  networks.
\newblock \emph{arXiv preprint arXiv:1903.12136}.

\bibitem[{Turc et~al.(2019)Turc, Chang, Lee, and Toutanova}]{turc2019well}
Iulia Turc, Ming-Wei Chang, Kenton Lee, and Kristina Toutanova. 2019.
\newblock Well-read students learn better: The impact of student initialization
  on knowledge distillation.
\newblock \emph{arXiv preprint arXiv:1908.08962}.

\bibitem[{Voita et~al.(2019)Voita, Talbot, Moiseev, Sennrich, and
  Titov}]{voita2019analyzing}
Elena Voita, David Talbot, Fedor Moiseev, Rico Sennrich, and Ivan Titov. 2019.
\newblock Analyzing multi-head self-attention: Specialized heads do the heavy
  lifting, the rest can be pruned.
\newblock In \emph{Proceedings of ACL}.

\bibitem[{Wang et~al.(2019{\natexlab{a}})Wang, Singh, Michael, Hill, Levy, and
  Bowman}]{wang2019glue}
Alex Wang, Amanpreet Singh, Julian Michael, Felix Hill, Omer Levy, and
  Samuel~R. Bowman. 2019{\natexlab{a}}.
\newblock {GLUE}: A multi-task benchmark and analysis platform for natural
  language understanding.
\newblock In \emph{Proceedings of ICLR.}

\bibitem[{Wang et~al.(2019{\natexlab{b}})Wang, Wohlwend, and
  Lei}]{wang2019structured}
Ziheng Wang, Jeremy Wohlwend, and Tao Lei. 2019{\natexlab{b}}.
\newblock Structured pruning of large language models.
\newblock \emph{arXiv preprint arXiv:1910.04732}.

\bibitem[{Warstadt et~al.(2018)Warstadt, Singh, and
  Bowman}]{warstadt2018neural}
Alex Warstadt, Amanpreet Singh, and Samuel~R. Bowman. 2018.
\newblock Neural network acceptability judgments.
\newblock \emph{arXiv preprint 1805.12471}.

\bibitem[{Williams et~al.(2018)Williams, Nangia, and
  Bowman}]{williams2018broad}
Adina Williams, Nikita Nangia, and Samuel~R. Bowman. 2018.
\newblock A broad-coverage challenge corpus for sentence understanding through
  inference.
\newblock In \emph{Proceedings of NAACL-HLT}.

\bibitem[{Yang et~al.(2019)Yang, Dai, Yang, Carbonell, Salakhutdinov, and
  Le}]{yang2019xlnet}
Zhilin Yang, Zihang Dai, Yiming Yang, Jaime Carbonell, Russ~R Salakhutdinov,
  and Quoc~V Le. 2019.
\newblock {XLNet}: Generalized autoregressive pretraining for language
  understanding.
\newblock In \emph{Proceedings of NeurIPS}.

\bibitem[{Zafrir et~al.(2019)Zafrir, Boudoukh, Izsak, and
  Wasserblat}]{zafrir2019q8bert}
Ofir Zafrir, Guy Boudoukh, Peter Izsak, and Moshe Wasserblat. 2019.
\newblock Q8{BERT}: Quantized 8bit {BERT}.
\newblock \emph{arXiv preprint arXiv:1910.06188}.

\end{thebibliography}
\bibliographystyle{acl_natbib}

\end{document}